\documentclass[adraft]{eptcs}
\usepackage[citations,hybrid]{markdown}

\usepackage[utf8]{inputenc}
\usepackage[T1]{fontenc}
\usepackage[margin=45pt]{geometry}

\title{Statistical relational learning and neuro-symbolic AI: what does first-order logic offer?}

 \author{Vaishak Belle\thanks{The material below is adapts discussions from a recent book by the author \cite{belle2023toward}, but has been extended to incorporate details and discussions about neuro-symbolic AI.}
\institute{University of Edinburgh, UK}
\email{vaishak@ed.ac.uk}}

\newcommand{\titlerunning}{SRL, NeSy and FOL}
\newcommand{\authorrunning}{Vaishak Belle}

\hypersetup{
  bookmarksnumbered,
  pdftitle    = {\titlerunning},
  pdfauthor   = {\authorrunning},
  pdfsubject  = {EPTCS},               
}

\begin{document}
\maketitle

\begin{abstract}
In this paper, our aim is to briefly survey and articulate the logical and philosophical foundations of using (first-order) logic to represent (probabilistic) knowledge in a non-technical fashion. Our motivation is three fold. First, for machine learning researchers unaware of why the research community cares about relational representations, this article can serve as a gentle introduction. Second, for logical experts who are newcomers to the learning area, such an article can help in navigating the differences between finite vs infinite, and subjective probabilities vs random-world semantics. Finally, for researchers from statistical relational learning  and neuro-symbolic AI, who are usually embedded in finite worlds with subjective probabilities, appreciating what infinite domains and random-world semantics brings to the table is of utmost theoretical import. 
\end{abstract}

\hypertarget{introduction}{%
\section{Introduction}\label{introduction}}

Statistical relational learning and neuro-symbolic AI are having a
growing impact on recent discussions regarding the connection between
(deep) learning and logical syntax
\cite{hitzler22neu,manhaeve18dee,d2020neurosymbolic,belle21log}.
These fields are motivated by the use of relations and logical
reasoning, specifically entailment and satisfaction, in learning
paradigms. They recognize that the world is inherently relational;
objects have properties and attributes, employees have employers that
are part of organisations, individuals are part of social networks. More
elaborately, in biological networks, genes are the fundamental units of
heredity that determine the traits of an organism. The process by which
genes are expressed in proteins involves several steps, including
transcription and translation, that are critical for the proper
functioning of cells and organisms. If we need to reason about
biological data, then algorithms that harness these relationships and
processes are fundamental.

Logical reasoning is crucial to deduce properties about these
relationships. For example, if we know that a dog is a mammal and a cat
is a mammal, we should be able to deduce that they both give birth to
live young, except unless those mammals that are monotremes. On the
other hand, if we know that insects cannot be mammals, then we can also
deduce that a dog cannot be an insect.

In the insular world of (deep) machine learning, where we assume that
objects are atomic and independent, the incorporation of logical
artifacts provides a much richer understanding of the world and its
objects, leading to a more nuanced and accurate interpretation of the
data.

A second important area where the impact of symbols extends to is the
notion of explainability in artificial intelligence. While traditionally
this area has been about interpreting decision boundaries, it now
includes attempts to use logical and programmatic objects to
characterize the prediction model. See, for example, recent celebrated
successes on deep concept induction \cite{lake15hum,ellis16sam}.
There is the more complex area of constructing mental models of the
users and providing explanations relative to that putative user
\cite{kambhampati20cha}. This area is also governed by symbolic and
formal principles. Symbolic principles such as sufficiency and necessity
are also being used to explore causal dimensions for explanations
\cite{chockler2021explanations}.

Despite this success and the many syntactic flavors explored in the
research community, the underlying logical language is essentially
propositional. Almost always, the outputs of the learning model (such as
a neural network) are either interpreted probabilistically over a finite
set of propositional variables, or as the numeric truth value accorded
to a finite set of propositions (in fuzzy logical approaches). Even in a
(probabilistic) logic programming setting such as ProbLog, cycle
breaking strategies are considered to reduce query-driven reasoning to
probabilistic labels over finitely many propositions via weighted model
counting \cite{sang05per,fierens11inf}.

In this paper, \emph{our aim is to briefly survey and articulate the
logical and philosophical foundations of using (first-order) logic to
represent (probabilistic) knowledge in a non-technical fashion}. Our
motivation is three fold. First, for machine learning researchers
unaware of why the research community cares about relational
representations, this article can serve as a gentle introduction.
Second, for logical experts who are newcomers to the learning area, such
an article can help in navigating the differences between finite vs
infinite, and subjective probabilities vs random-world semantics.
Finally, for researchers from statistical relational learning (StarAI)
and neuro-symbolic AI (NeSyAI), who are usually embedded in finite
worlds with subjective probabilities, appreciating what infinite domains
and random-world semantics brings to the table is of utmost theoretical
import: in fact, one may even argue it is needed to make the area
well-defined! StarAI and NeSyAI wishes to generalize to unforeseen
examples, and deal with statistical properties, and yet their prove
properties for finite models and degrees of belief.

Of course, the use of a relational language (with a finite domain
notwithstanding), that only serves as syntactic sugar for the
specification for probabilistic models, is not without its advantages.
It can greatly simply the specification of relations and hierarchies
over random variables, and can also speed up probabilistic inference in
some cases by appealing to first-order resolution
\cite{van-den-broeck13lif}. \emph{But by incorporating probabilistic
reasoning into first-order logic, we might be able to better deal with
infinite domains and incomplete knowledge in a more principled way, and
allowing for the specification of random properties (and not simply
subjective probabilities, as assumed by current mainstream languages) in
an open-world setting.} However, there will still challenges to be
addressed, such as the computational complexity of reasoning with large
knowledge bases and the need for effective methods of learning from
data.

\hypertarget{why-representation-matters}{%
\section{Why representation matters}\label{why-representation-matters}}

It is worth reflecting on the nature of knowledge representation and how
it affects modeling and learning. Regardless of whether we use a logical
representation (such as a logic program), a probabilistic one (such as a
Bayesian network), or a connectionist one (such as a neural network),
our goal is to represent information and capture the knowledge of a
robot or vision system. We want to understand how this representation
produces interesting conclusions that determine what the robot knows and
understands about the world.

While acquisition is undoubtedly important, for now, we are focusing on
the language used for representing information. It is crucial to equip
this language with tools for modeling and reasoning about the world. In
a statement remarkably similar in spirit, Pearl writes
\cite{pearlthe}:

\emph{This is why you will find me emphasizing and reemphasizing
notation, language, vocabulary and grammar. For example, I obsess over
whether we can express a certain claim in a given language and whether
one claim follows rom others. My emphasis on language also comes from a
deep conviction that language shapes our thoughts. You cannot answer a
question that you cannot ask, and cannot ask a question that you have no
words for.}

Which language should we choose, and what should we represent? Turing
machines are foundational structures for investigating the computable
universe. They offer minimal, yet powerful read and write operations.
However, being a computational abstraction means we have no means to
concretely talk about what the robot knows about the environment it
operates in. Likewise, binary system code runs robotic software, but is
too coarse and opaque to explain a purposeful agent. Vectorized
representations and non-linear decision boundaries as obtained by
sigmoid neural units over many interconnected layers are also too coarse
to explain the functioning of a deep network. Although these models seem
capable of capturing a potentially infinite hypothesis space and
high-dimensional data, including high-resolution images such as
mammograms, an operational description is too opaque for humans to
decipher.

Therefore, we require a modeling framework that is understandable to
domain experts and allows for the updating and provision of new
information, while also being suitable for computational processes. In
particular, by a representation, we mean a mapping from one entity to
another. The former is concrete and computationally processable. For
example, ``rain'' might stand for the assertion that it is raining
today. Reasoning is the computational processing of representations to
produce new knowledge. We focus on formal (symbolic) logic, where all
information is stored explicitly or implicitly using symbols.

The distinction between explicit and implicit knowledge is important:
explicit knowledge is often the information that is directly modelled or
provided by the domain expert. This could come in the form of rules,
databases, knowledge bases, graphs, or any other structured data.
Implicit knowledge is what is obtained from the explicit knowledge
through one or more reasoning steps.

Reasoning can be understood in two ways. First, it can define the
mathematical framework for obtaining explicit knowledge from implicit
knowledge. In logic, we often provide a semantics and/or proof theory
that determines when sentences are ``entailed'' from a knowledge base (a
set of sentences). Second, reasoning can provide an implementation
strategy for computing entailments: an algorithm that takes a knowledge
base and a query as input and outputs whether the query is entailed.
Standard platforms for designing such algorithms include theorem
provers, model checkers, and satisfiability solvers.

\hypertarget{logic-and-probability-how-did-we-get-here}{%
\section{Logic and probability: how did we get
here?}\label{logic-and-probability-how-did-we-get-here}}

Leibniz perhaps first suggested that thinking could be understood as
symbol manipulation, but it was only with the results of Boole and Frege
that an algebraic view of logic became possible. Symbolic logic was
primarily used for studying the foundations of mathematics until the
19th century. David Hilbert's program, Kurt Gödel's seminal
incompleteness theorem, and other such inquiries studied the arithmetic
of natural numbers and what logically follows from that arithmetic. John
McCarthy, among others, began to see first-order logic as a formalism
for capturing commonsense and language only since the 1960s
\cite{mccarthy68pro,mccarthy68sit,morgenstern11joh}. He suggested
that the system's knowledge could be represented in a formal language,
and a general-purpose algorithm could then determine the necessary
actions to solve the problem at hand. A major advantage of this approach
is that the representation can be scrutinized and understood by external
observers, allowing them to improve the system's behavior by making
statements to it.

Numerous such languages emerged in the years to follow, but first-order
logic remained at the forefront as a general and powerful option.
Unfortunately, one of the main arguments against a logical approach is
the pervasive uncertainty present in almost every domain of interest.
This uncertainty can manifest in various forms, such as measurement
errors (e.g.~readings from a thermometer), the absence of categorical
assertions (e.g.~smoking may be a factor for cancer, but cancer is not
an absolute consequence for smokers), and the presence of numerous
``latent'' factors. These factors include causes that the modeler may
not have taken into account, which question the legitimacy of the model.

Modeling uncertain worlds required a rigorous formulation, which has
been achieved through probabilistic models, of course. Probabilistic
models, obtained either by explicit specification or implicitly induced
by modern machine learning methods such as deep learning, have
supercharged the application of statistical methods in language
understanding, vision, and data analysis more generally.

Despite the success of probabilistic models, we observe that they are
essentially propositional, yet are deployed in an inherently relational
world. That is, they easily make sense of ``flat'' data, where atomic
events are treated as independent and identically distributed random
variables. This realization has led to the design of probabilistic
concept languages, culminating in the area of statistical relational
learning, neuro-symbolic AI, and other hybrid languages.

Note that there is a considerable history on the unification of logic
and probability, including early work by Carnap and Gaifman
\cite{carnap51log,gaifman64con}. Interested readers can refer to
\cite{belle21log} for a discussion on the historical development.

\hypertarget{first-order-logic}{%
\section{First-order logic}\label{first-order-logic}}

First-order logic is, as discussed above, a widely studied formalism. To
contrast it with popular proposals in statistical learning, let us start
with an example often used in that community, which involves smoking
within social networks. We will then discuss variations allowed by
propositional and first-order logic, while also briefly demonstrating
their features. As will become clear, virtually every formalism from
that community does not support arbitrary first-order logic. There is
one good reason for this: first-order logic is only semi-decidable, and
computing entailments is not tractable \cite{boerger97the}.
(Although first-order logic with reals as its domain is decidable, it is
best used to reason about arithmetic properties, as seen in
satisfiability modulo solvers.)

Let us imagine two individuals \(a\) (for Alice) and \(b\) (for Bob)
such that \(a\) is a smoker and the two are friends. Using \(S(x)\) as a
predicate to mean \(x\) is a smoker and \(F(x,y)\) to mean \(x\) is a
friend of \(y\), we have: \(S(a) \land F(a,b)\). There are two
propositional assertions here, connected using a conjunction, to say
both are true in the knowledge base (KB).

Suppose we found that smokers influence those in their social networks
to also smoke, in which case, we would wish to add \(S(b)\) to the KB.
However, rather than working this out by hand, we might wish to express
this observation using an implication rule:
\(S(a) \land F(a,b) \supset S(b)\). Such rules involve only constants,
and so, with quantifiers, we will be able to say that the smokers rule
applies to every individual:
\(\forall x,y.~S(x) \land F(x,y) \supset S(y)\). The quantification is
over a ``domain of discourse'', which defines the constants to which the
expression following the quantification applies.

Naturally, the knowledge base (KB) contains ``facts'', i.e., ground
atoms, such as \(S(a)\) and \(F(a,b)\), in addition to quantified
formulas. These facts constitute explicit knowledge. By applying these
facts to the rules, we can derive \(S(b)\), which is implicit in the
sense of the terminology used earlier. In this way, the truth value for
every instance of \(S(x)\) can be determined from explicit or implicit
knowledge, similar to a database. This denotes the case of ``complete
knowledge.''

Incompleteness is usually expressed using disjunctions and existential
quantifiers. For example, assuming a third constant \(c\) for Carol:
\((S(a) \lor S(c)) \land \neg (S(a) \land S(c))\) says that either Alice
or Carol is a smoker but not both.

The use of an existential statement can declare the existence of an
individual with a certain property, but not specify who that individual
might be. For example, the following sentence states that there is a
friend of Alice who is not Carol:
\((\exists x.~F(x,b)) \land \neg F(c,b)\). Suppose we further stipulate
that no one can be their own friend: \(\forall x.~\neg F(x,x)\). Of
course, if our domain were finite and consisted only of the individuals
discussed so far - \(\{a,b,c\}\) - then \(F(a,b)\) would follow from the
above two sentences. However, if the domain is infinite, then infinitely
many candidates are possible friends of Bob, but we are not certain of
any one of them.

\hypertarget{infinite-domains}{%
\subsection{Infinite domains}\label{infinite-domains}}

In addition to incomplete information about domains, other types of
incomplete knowledge can also be expressed with infinite domains. For
instance, when quantifiers range over an infinite set, the sentence
\(\forall x.~x\neq a \supset S(x)\) states that there are infinitely
many individuals who are smokers, except for Alice who might or might
not be a smoker. Therefore, it models unknown atoms. To represent
unknown values, we can use the following sentence:
\(\forall x(S(x) \supset eyecolor(x) \neq green )\), which declares that
among smokers in our population, of whom there may be infinitely many,
none has green eyes.

Although there are a wide range of formalisms in statistical relational
learning, constants are typically fixed in advance and are finite.
Functions are rarely allowed, and often the interpretation of predicates
is fixed as well: that is, the set of ground atoms is fixed, as well as
their truth values. When these formalisms allow atoms to be decorated
with probabilities, there is some flexibility in terms of the
interpretation of predicates, in the sense that a positive instance is
more likely than a negative instance, or vice versa.

As noted, the full first-order language, especially over a possibly
infinite domain, can allow for unknown sets of atoms, their possible
values, and varying sets of objects (e.g., when the set of humans is
left open). This makes standard first-order logic a powerful and
general-purpose language for representing knowledge, and we believe it
can be a promising candidate for the next generation of probabilistic
relational models.

In fact, keeping the domain undefined is often referred to as
``open-world'' modelling. Note that closeness is not difficult to
implement in open-world models, should we so wish. For example, the
sentence below says that Alice and Bob are both smokers, and everyone
else is not:
\(\forall x[ ((x=a\lor x = b) \supset S(x)) \land ((x\neq a \land x\neq b) \supset \neg S(x))].\)

\hypertarget{probabilities-on-formulas}{%
\subsection{Probabilities on formulas}\label{probabilities-on-formulas}}

The standard way to represent incomplete knowledge in symbolic logic is
through the use of disjunctions and existential quantifiers. A logical
disjunction expresses qualitative uncertainty, but does not commit to
which of the disjuncts is more likely than the other.

For example, suppose we toss a coin and find that it lands on heads nine
out of ten times. We might believe it is a loaded coin and ascribe a
probability of 0.9 to observing heads on the next toss. Suppose we read
three plausible reports, two of which suggest that Alice is not a
smoker, but the third asserts that she is. We might ascribe a
probability of 2/3 to \(S(a)\). However, given that a report about Alice
being a smoker might have been more recent or that Alice accidentally
admitted she occasionally smokes, we might treat it as a true but
lesser-known fact and ignore the other reports. Whichever the case may
be, we need a language to reason about the \emph{degrees of belief} --
i.e., subjective probabilities -- of the agent.

One of the simplest and most widely studied approaches to dealing with
degrees of belief in logic involves the concept of \emph{possible
worlds}, where each world represents a distinct interpretation of the
underlying knowledge base. We ascribe a probability (or more generally,
a weight) to each world. To determine the probability of a statement
\(S(a)\), we sum the probabilities of all worlds where \(S(a)\) is true.
We may also use weights and consider the ratio of that sum to a
\emph{normalization factor}, which is simply the sum of the
probabilities of all worlds.

\hypertarget{probabilities-on-atoms}{%
\subsection{Probabilities on atoms}\label{probabilities-on-atoms}}

Suppose \(S(a)\) is believed with a probability of 1. What is the
probability that Bob is also a smoker? Given that \(S(a)\) is entailed
by KB, meaning it holds in every possible model of the KB, we cannot
infer anything about \(S(b)\) without further information. The
probability of \(S(b)\) depends on the possible worlds. If \(S(b)\) is
also entailed by the KB, it would be accorded a probability of 1 as
well. However, if some models of the KB did not satisfy \(S(b)\), then
the probability of \(\neg S(b)\) can be obtained by summing the
probabilities of all such worlds, say this is \(r\). The probability of
\(S(b)\) would then be \((1-r)\). If the worlds all have equal weights,
and \(S(b)\) is false in \(r\) out of \(n\) worlds, then the probability
of \(S(b)\) would be \((n-r)/n\). From a logical viewpoint, the
extension of the predicate \(S\) is different in every world, and it so
happens that \(a\) is in every such extension while \(b\) is only in
some.

Suppose we were able to assign a probability of 4/5 to Alice and Bob
being smokers based on the deterministic rule mentioned above. This rule
can be represented in the KB as
\(4/5\colon~ (S(a) \land F(a,b)) \supset S(b)\), indicating that in 4
out of 5 worlds, Alice influences her friend to smoke. If all worlds had
equal weights, this would mean that Alice influences her friend to smoke
in 4/5 of those worlds.

To further understand the rule, let's consider two scenarios with
different probabilities for Alice being a smoker. Assume
\(\{S(a), F(a,b)\}\) is believed with probability 1. The noisy rule
states that \(S(b)\) is true in four-fifths of the worlds, giving us a
probability of 4/5 for \(S(b)\).

Now, let's assume \(F(a,b)\) is believed with probability 1, but
\(S(a)\) only with a probability of 2/3. Assuming equal weights to
worlds for ease of explanation, \(S(a)\) holds in two-thirds of the
worlds, and in four-fifths of those worlds, the noisy rule requires that
\(S(b)\) holds. Therefore, the probability of \(S(b)\) is
\(2/3 \times 4/5 = 8/15.\)

Let's consider a simple example where a unique distribution would not
arise. Suppose \(\forall x, y(F(x,y))\) is believed with probability 1.
This means that every individual is a friend of everybody else
(including themselves). Now suppose \(S(a) \lor S(c)\) is believed with
probability 1. In this case, no specification is provided for the
individual probabilities of \(S(a)\) and \(S(c)\). It is not acceptable
to assign a probability of 0 to both \(S(a)\) and \(S(c)\). However,
provided that this requirement is satisfied, there are many
possibilities. Most statistical relational languages will either
disallow such specifications or make assumptions about how the
distribution should be defined for such disjunctions. Notable exceptions
allowing for imprecise probabilities do exist, but they are often not in
a first-order formalism. See \cite{bacchus99rea,belle17rea} for
examples that involve first-order logic and multiple distributions.

Suppose we are interested in the probability of \(S(b)\) given the noisy
smokers rule above and a similar instance resulting from \(S(c)\):
\(4/5\colon~ (S(c) \land F(c,b)) \supset S(b)\). It is not hard to see
that the probability of \(S(b)\) should be 4/5. This is because every
world should satisfy \(S(a)\) or \(S(c)\) or both. Due to the universal
rule about everyone being friends, \(F(a,c)\) and \(F(b,c)\) both hold.
Consequently, in four-fifths of the worlds (assuming equal weights for
simplicity), the smokers rule would imply that \(S(b)\) must hold.
Therefore, its probability is also 4/5.

Suppose we instead said \(S(a) \lor S(c)\) with a probability of 1/5. In
one-fifth of the possible worlds, \(S(a)\) or \(S(c)\) or both would be
true. In four-fifths of those worlds, the smokers rule would mean that
\(S(b)\) must be true as well. Therefore, the probability of \(S(b)\)
would be \(4/5 \times 1/5 = 4/25\) assuming equal weights.

\hypertarget{probabilities-on-quantified-formulas}{%
\subsection{Probabilities on quantified
formulas}\label{probabilities-on-quantified-formulas}}

The examples from the previous section could be expressed more
succinctly with the following noisy quantified rule:
\(4/5\colon~ \forall x, y(S(x) \land F(x,y) \supset S(y)).\) This rule
states that in four-fifths of the worlds (assuming equal weights to
worlds), friends of smokers are smokers too.

Suppose we know for certain that Alice and Bob are friends, and we
ascribe a probability of 2/3 to the atom \(S(a)\). Then, in two-thirds
of the worlds, \(S(a)\) holds. In four-fifths of these worlds, the
universal rule, together with the certain knowledge that \(F(a,b)\),
means that \(S(b)\) holds. The probability of \(S(b)\) is then
\(2/3 \times 4/5 = 8/15.\)

However, one might find it striking how the reading of this rule
contrasts with the statistical observation that motivates it.
Presumably, we would like to say that in 80\% of the population, friends
of smokers are also smokers. This is statistical information, of course,
but the way the rule is written requires that the rule must hold in
four-fifths of the worlds. In other words, if in every world there is
some smoker whose friend is not a smoker, then the rule clearly holds in
no world, and so its probability should be 0.

This is an early observation in the field of probabilistic logics
\cite{abadi94dec,bacchus94for,bacchus96fro,halpern03rea,halpern90an}.
The possible-worlds model with probabilities over worlds is used to
define degrees of belief, but it is not appropriate for statistical
information. To represent statistical information, a different formal
setup is required. This setup, called the \emph{random-worlds}
semantics, allows probabilities to be ascribed to domain elements. With
this approach, we can reason about the probability of a randomly chosen
person being a smoker, or the probability of two randomly chosen
individuals being friends, and so on. This would be useful in cases
where the smokers universal is known to not be true in the real world.

In most statistical relational and neuro-symbolic languages, the
intended statistical interpretation is often ignored, resulting in noisy
quantified rules. These rules are commonly interpreted using the
possible-worlds model.

Likewise, probabilities on existentially quantified formulas are to be
understood the same way. For example: \(1/5\colon~\exists x(S(x))\) says
that in one-fifth of the worlds (assuming equal weights on worlds),
there is someone who is a smoker. It is \emph{not} saying that one out
of five individuals is a smoker or that there is a 20\% chance that a
randomly chosen individual is a smoker.

Both of these topics require a random-worlds semantics, and integrating
that with a possible-worlds semantics could be an exciting area of
research for mainstream statistical relational and neuro-symbolic AI.

\hypertarget{what-lies-beneath}{%
\subsection{What lies beneath}\label{what-lies-beneath}}

ProbLog, probabilistic databases, Markov logic networks, Bayesian
networks, factor graphs \cite{raedt16sta}, and their neuro-symbolic
extensions can be understood either in terms of a computational
framework called \emph{weighted model counting}
\cite{chavira08on,sang05per}, or via a fuzzy semantics that
accords non-binary truth values to atoms. Let us discuss them in turn.

Given a logical language with n propositions, and therefore, \(2^n\)
possible truth assignments (or worlds)
\({\mathcal M} = \{M\_1, \ldots, M\_{2^n} \}\), consider a weight
function \(\mu\) mapping literals to non-negative reals. By extension,
let the weight of a world \(\mu(M)\) be defined as the product of the
weights of the literals that hold at \(M\). Then the weighted model
count of a propositional formula \(\phi\) is defined as:

\(\displaystyle {\it WMC}(\phi) \doteq {\sum\_{\{M \in {\mathcal M},~ M\models \phi\}}\mu(M)}\bigg/ {\sum\_{\{M \in {\mathcal M}\}}\mu(M)}.\)

A Bayesian network can be encoded as a propositional knowledge base,
with the conditional dependences expressed as implications
\cite{chavira08on,sang05per}. In Markov logic networks,
propositions appearing in a clause correspond to the dependent random
variables in an undirected graphical model. The probability function of
the nodes can be further expressed in terms of a weight function of the
above sort. In this case, the probability of some query \(q\) (a
propositional formula) is obtained using:

\({{\it WMC}({\it KB} \land q)} \bigg/ {{\it WMC}({\it KB})}.\)

In other words, relative to the models of KB, we compute the weights of
those worlds additionally satisfying \(q\).

There are a few notable exceptions to picture in the statistical
relational learning community; see \cite{belle17ope} for more
discussions. For example, in BLOG \cite{milch05blo}, function
symbols are allowed in the language over an infinite domain, thus
supporting open-world modeling. However, for computational reasons,
various restrictions are placed both for the language and its
implementation. Well-defined programs, for example, restrict the use of
logical connectives. When computing probabilities, it avoids
instantiating infinitely many terms by sampling values for function
symbols. When it comes to termination, moreover, a number of structural
conditions need to hold, including one that the quantifiers over
formulas can only range over finite sets.

In a second approach to the semantics of neural learning in logic,
binary truth is relaxed. In particular, because the outputs of neural
networks are real-valued and need to be interpreted logically, there has
been a recent revival of fuzzy logic in neuro-symbolic AI
\cite{besold17neu,garcez02neu,lamb20gra}. Fuzzy logic allows for
degrees of truth, unlike classical logic which assumes a proposition is
either true or false. In fuzzy logic, logical connectives are
interpreted as maximum and product operators, where the maximum operator
corresponds to disjunction and the product operator corresponds to
conjunction.

When the inputs to these operators are either 0 or 1, it can be seen
that maximum and product collapses to the classcal interpretation.
However, when the inputs are reals between 0 and 1, formulas are
evaluated to values that do not correspond to being strictly true or
false. This makes the semantics of complex formulas hard to interpret.
Although there is some research on first-order fuzzy logic over
non-finite domains, much of the work on neuro-symbolic AI is limited to
finite domains \cite{garcez02neu}.

A common argument in favor of fuzzy logic is that it may be more
feasible computationally \cite{krieken20ana,krieken22ana}, although
this comes at the cost of a hard-to-interpret semantics. However, recent
discussions have raised the point that neural networks are trained for
conditional probabilities, making it reasonable to interpret the output
in a probabilistic manner \cite{krieken20ana,krieken22ana}. As a
result, it may be more appropriate to consider using probabilistic
(logical) languages. Striking a balance between ease of computation and
interpretability of semantics is crucial, and perhaps the final decision
may depend on the specific application.

\hypertarget{outlook-and-conclusions}{%
\section{Outlook and conclusions}\label{outlook-and-conclusions}}

In the above sections, we discussed the importance of reasoning using
the friends-smokers example. We also explored the advantages of
first-order logic in terms of the expressiveness it offers for
open-world scenarios. However, we did not delve into the strategies that
allow us to reason in a decidable way with fragments of first-order
logic. Fortunately, there are numerous methods available to tackle this
issue, and we shall mention them in brief.

As is well known, areas such as descriptive complexity and finite model
theory demonstrate that logic could serve as an abstraction for
understanding computation vis-à-vis representation. It is also
well-known that there is a trade-off between the expressiveness of the
language and the tractability of the reasoning task.

On the one hand it is widely acknowledged that relations and more
generally, the expressiveness of first-order logic is extremely useful
for capturing language and concepts required for common-sense reasoning.
This explains the use of first-order logic in database theory,
statistical relational learning, automated planning, knowledge
representation (e.g., description logics, and reasoning about actions)
and verification. But at the same time the entailment problem is only
semi-decidable.

Since Goedel's celebrated results on the intractability of logic, there
has been a wide range of approaches to deal with this trade-off
especially in the knowledge representation, symbolic computation and the
statistical relational learning communities. They can be broadly
classified into three categories. The easiest approach is to simply
restrict the language to propositional logic. Popular frameworks in
statistical relational learning such as Markov logic networks and
relational Bayesian networks as well as relational probabilistic
planning languages take this approach
\cite{corander13lea,kok10lea,lowd13lea,younes04ppd}.

A second approach is to limit the expressiveness of the language in
terms of the size of the predicates (but not necessarily in terms of the
size of the domain). Description logics are popular concept
representation languages in knowledge representation that only allow
predicates with utmost two arguments. When the domain is further assumed
to be finite, certain other computational tasks beyond entailment turn
out to be tractable \cite{calvanese13dat,koller97p-c}.

A third approach, stemming from the knowledge representation community
but less popular in other communities, is to limit reasoning. Here, the
entailment relation itself is weakened by way of introducing
non-traditional semantics \cite{liu05tra}. However, because the
semantical apparatus is often complicated, they have understandably not
been pursued by or used in other communities.

Somewhere in between the three approaches, there is also the well-known
Horn clause restriction, seen most prominently in logic programming and
its derivatives. It potentially allows for an infinite Herbrand
universe, but the grounding of the program is only performed against a
ground query, which permits a propositional encoding sometimes
\cite{fierens11inf}. This enables the use of logic programming in a
wide variety of applications, including statistical relational learning
by decorating clauses further with probabilities. But most commong
flavors are limited in only allowing one positive literal in a clause,
and is non-monotonic.

A fourth approach is to exploit symmetries in first-order
representations. For example, the area of lifted reasoning avoids
grounding a probabilistic relational model by appealing to first-order
resolution \cite{van-den-broeck13lif}. In somewhat a similar spirit,
a recent result shows that there is an intuitive fragment of first-order
disjunctive knowledge, for which reasoning is decidable and can be
reduced to propositional satisfiability \cite{belle19imp}. Knowledge
bases in this fragment essentially amounts to universally quantified
first-order clauses, but without arity restrictions and without
restrictions on the appearance of negation. Queries, however, are
expected to be ground formulas. They achieve this result is by showing
how the entailment over infinitely many infinite-sized structures can be
reduced to a search over finitely many finite-size structures. The crux
of the argument lies in showing that names not mentioned in the
knowledge base and the query behave identically (in a suitable formal
sense).

Overall, if we are extend the statistical relational and neuro-symbolic
formalisms to full first-order expressiveness, some of these strategies
need to be combined and extended carefully, and this is an exciting area
for future research.

\end{document}